\documentclass[journal]{IEEEtai}

\usepackage[colorlinks,urlcolor=blue,linkcolor=blue,citecolor=blue]{hyperref}

\usepackage{color,array}
\usepackage[utf8]{inputenc} 
\usepackage[T1]{fontenc}    
\usepackage{hyperref}       
\usepackage{url}            
\usepackage{booktabs}       
\usepackage{amsfonts}       
\usepackage{nicefrac}       
\usepackage{microtype}      
\usepackage{lipsum}
\usepackage{fancyhdr}       
\usepackage{graphicx} 
\usepackage{pifont}
\usepackage{amsmath}
\graphicspath{{figures/}}     
\usepackage{multirow}
\newcommand{\greentick}{\textcolor{green!50!black}{\ding{51}}}
\newcommand{\redcross}{\textcolor{red}{\ding{55}}}
\newtheorem{definition}{\bf{Definition}}[section] 
\usepackage{wrapfig}
\usepackage[dvipsnames]{xcolor} 

\definecolor{softblue}{RGB}{100, 130, 200}
\definecolor{softgreen}{RGB}{100, 180, 100}
\definecolor{softorange}{RGB}{230, 150, 80}
\definecolor{softpurple}{RGB}{160, 120, 200}
\definecolor{softgray}{gray}{0.4}

\begin{document}

\title{A Survey of Graph Unlearning}

\author{Anwar Said, Ngoc N. Tran, Yuying Zhao, Tyler Derr, Mudassir Shabbir, Waseem Abbas, Xenofon Koutsoukos

\thanks{
\noindent{This material is based upon work supported by the National Science Foundation under Grant 2325416 and Grant 2325417. }\\
{Anwar Said, Ngoc N. Tran, Yuying Zhao, Tyler Derr, and Xenofon Koutsoukos are with the Computer Science Department, Vanderbilt University, Nashville, TN, USA.} \\
{Mudassir Shabbir is with the Computer Science Department, Information Technology University, Lahore, Pakistan.}\\
{Waseem Abbas is with the Systems Engineering Department, University of Texas at Dallas, Richardson, TX, USA.} \\
{Anwar Said and Ngoc N. Tran contributed equally to this work.}\\
}}


\maketitle

\begin{abstract}
Graph unlearning emerges as a crucial advancement in the pursuit of responsible AI, providing the means to remove sensitive data traces from trained models, thereby upholding the \textit{right to be forgotten}. It is evident that graph machine learning exhibits sensitivity to data privacy and adversarial attacks, necessitating the application of graph unlearning techniques to address these concerns effectively. In this comprehensive survey paper, we present the first systematic review of graph unlearning approaches, encompassing a diverse array of methodologies and offering a detailed taxonomy and up-to-date literature overview to facilitate the understanding of researchers new to this field. To ensure clarity, we provide lucid explanations of the fundamental concepts and evaluation measures used in graph unlearning, catering to a broader audience with varying levels of expertise. Delving into potential applications, we explore the versatility of graph unlearning across various domains, including but not limited to social networks, adversarial settings, recommender systems, and resource-constrained environments like the Internet of Things, illustrating its potential impact in safeguarding data privacy and enhancing AI systems' robustness. Finally, we shed light on promising research directions, encouraging further progress and innovation within the domain of graph unlearning. By laying a solid foundation and fostering continued progress, this survey seeks to inspire researchers to further advance the field of graph unlearning, thereby instilling confidence in the ethical growth of AI systems and reinforcing the responsible application of machine learning techniques in various domains.
\end{abstract}

\begin{IEEEkeywords}
Adversarial Attacks, Data Privacy, Foundation Models, Graph Representation Learning, Graph Unlearning, Machine Unlearning, Recommender Systems
\end{IEEEkeywords}

\section{Introduction}
\label{sec:introduction}

In the ever-evolving landscape of artificial intelligence (AI) and data-driven decision-making, machine learning (ML) has emerged as a transformative force, empowering systems to learn from vast amounts of information and make intelligent predictions. However, as our digital world evolves, so do the ethical and privacy challenges it presents \cite{mitchell2006discipline}. Enter the pioneering concept of machine unlearning—an awe-inspiring technological frontier designed to harmonize the power of ML with individual data rights and data privacy regulations \cite{bourtoule2021machine,cao2015towards}.
These policies allow us to gracefully erase the footprints of sensitive data from trained models, which has been proven to be easily recoverable through various adversarial attacks~\cite{zhao2021feasibility,jayaraman2022attributeinferenceattacksjust,olatunji2021membership,zhang2022inference}.
Machine unlearning represents an extraordinary leap forward in the pursuit of responsible AI, empowering individuals to exercise their \emph{right to be forgotten}~\cite{rosen2012right,regulation2018art,CCPA2018}, and instilling confidence in AI's ethical evolution~\cite{10.1145/3696410.3714740}.

An area where this ethical challenge becomes especially relevant is graph machine learning (GML).
%
Unlearning is an important technique in almost every application where GML models are utilized, including social networks, healthcare systems, financial networks, transportation networks, and more.
As an example of unlearning, consider a company that has trained a GML model on a social network data to suggest potential new friends based on user interactions and their features. If a user requests their data be removed under privacy regulations, the model must undergo unlearning to eliminate the impact of their data \cite{bourtoule2021machine}.
Alternatively, from a security perspective, unlearning plays an essential role in addressing challenges and threats posed by adversarial attacks, which often involve malicious manipulations of nodes, edges, or features to compromise the functionality of GML models. In scenarios where adversaries inject harmful data into the graph structure--commonly referred to as poisoning attacks, it becomes crucial to effectively unlearn the malicious data without compromising benign portions of the graph or reducing the model's predictive accuracy. Unlearning adversarial perturbations ensures the system’s integrity while safeguarding it from exploitation, maintaining its usability in critical applications such as recommendation systems, fraud detection, and network security~\cite{wu2024graphmu}.

Though unlearning GML models have a great potential, the complexity of graph-structured data poses a challenge for unlearning in graphs. Relationships between nodes and edges are intricate in graphs, forming a dense web of interconnected information \cite{dukler2023safe}. Similarly, GML uses message passing to aggregate information from neighboring nodes and edges. A single removed node, edge, or subgraph can influence the embeddings of distant nodes through multiple iterations of message passing. This makes it challenging to completely erase the influence of the removed data without retraining the model from scratch. Thus, to unravel the impact of forgotten data while conserving the graph's structure, sophisticated algorithms are required. These algorithms must be able to untangle these intricate connections without compromising the graph's functionality \cite{pan2023unlearning}.
Moreover, as data privacy regulations and the rights of individuals to control their data become increasingly important, the ability to comprehensively remove sensitive information from graph-structured data becomes a crucial ethical safeguard. Ensuring the effective and complete eradication of forgotten data from graphs gives individuals more control over their information, instills confidence in ML systems, and upholds ethical standards for the responsible management of data \cite{chen2022graph}.

Significant efforts have been made in the design of effective graph unlearning strategies, indicating a growing awareness of the importance of addressing data privacy and individual rights in the domain of GML~\cite{cheng2023gnndelete,bettini2018privacy}. These works have yielded notable advancements, demonstrating the researchers' and practitioners' commitment to tackling the complexities of graph unlearning. However, it is essential to recognize that the discipline is still relatively young and continues to evolve rapidly. As graph unlearning encounters the unique obstacles posed by interconnected networks, continued effort and creative solutions are required to overcome these obstacles.

In light of the rapidly changing landscape of graph unlearning and the urgent need to resolve data privacy concerns in interconnected networks, we present this comprehensive survey as a crucial step in gaining a better understanding of the field. This survey seeks to summarize, analyze, and classify existing graph unlearning techniques, shedding light on their merits, limitations, and practical implications. Specifically, we collected papers by first gathering influential works and related surveys as seeds. From these, we expanded our survey by including more recent works that refer to those foundational papers. By collecting and analyzing these strategies, we hope to provide researchers and practitioners with a clear map for navigating this new domain. In addition, we envision this survey as a catalyst for future progress, outlining potential research and development avenues that will influence the future of graph unlearning. We offer the following contributions. 

\begin{itemize}
    \item We propose a comprehensive taxonomy that categorizes graph unlearning approaches into two distinct groups: exact unlearning, and approximate unlearning, each exhibiting unique characteristics and diverse applications.
    \item We explore potential applications of graph unlearning across various fields, and delve into intricate details of their current development.
    \item We emphasize multiple open research directions that demand further investigation, driving the pursuit of novel frontiers in this captivating domain.
\end{itemize}

In recent years, a number of comprehensive survey papers have addressed the topic of machine unlearning~\cite{nguyen2022survey,qu2023learn,xu2023machine,qu2023learn,wang2024machine}.
Despite their valuable contributions, it is important to note that the aforementioned studies concentrate solely on general machine unlearning paradigms, which entails fundamentally different challenges and techniques than GML. Given that standard machine learning algorithms are not directly applicable to graph-structured data, unlearning methods designed for conventional settings cannot be straightforwardly transferred to graph learning scenarios. To the best of our knowledge, the only previous work that addressed \emph{graph unlearning} is OpenGU~\cite{fan2025opengu}; however, this survey focuses mostly on benchmarking recent results and does not adequately represent the generate landscape of graph unlearning. This survey aims to bridge this gap by presenting a systematic and focused review of the emerging field of graph unlearning.



\section{Preliminaries}
\label{sec:prelim}

In this section, we introduce several key concepts and definitions that will be used in the rest of the paper.

Generally, a graph is defined as $G = (V, E, X, X_e)$, where $V$ represents a set of nodes denoted by \(V = \{v_1, v_2, ..., v_n\}\), and \(E\) represents a set of edges denoted by \(E = \{(v_i, v_j)\}\), signifying pairwise associations between nodes. The node feature matrix, \(X \in \mathbb{R}^{|V| \times d}\), captures the attributes or characteristics of each node, where \(d\) represents the dimensionality of node features. The edge feature matrix, \(X_e \in \mathbb{R}^{|E| \times d^\prime}\), is optional and can be empty (\(X_e = \emptyset\)) in unweighted graphs. Here $d^\prime$ is the dimensionality of edge features \cite{jegelka2022theory}. Lets define $n$ the number of nodes and $m$ the number of edges in $G$. We also define $\mathcal{Y} = \{y_1,y_2,..,y_n\}$ consists of the labels associated with each node in the graph. The length of $|\mathcal{Y}|$ might be less than $n$ in some cases where fewer node labels are available. Consider the set of graphs or a graph dataset denoted as $\mathcal{G} = \{G_1, G_2, ..., G_N\}$ and its corresponding label set $L =  \{l_1, l_2, \ldots, l_N\}$. Here $N$ denotes the number of samples (graphs) in the training data. Let $\mathcal{D}$ represent the corresponding embeddings (dataset) of $\mathcal{G}$, given as $\mathcal{D} \in \mathbb{R}^{N \times h}$ and obtained through a graph embedding function denoted as $f(.)$. 

Let us define \(\theta^o = \mathcal{A}(\mathcal{D})\) as the model parameters obtained by applying the learning algorithm \(\mathcal{A}\) to the full dataset \(\mathcal{D}\). Let \(\mathcal{S} \subseteq \mathcal{D}\) denote the forget set, and define \(\mathcal{D}^\prime = \mathcal{D} \setminus \mathcal{S}\) as the dataset obtained after the removal request. We denote by \(\theta^r = \mathcal{A}(\mathcal{D} \setminus \mathcal{S})\) the parameters of the retrained model obtained by applying \(\mathcal{A}\) to the reduced dataset \(\mathcal{D}^\prime\). Additionally, let \(\theta^u = \mathcal{U}(\theta^o, \mathcal{S}, \mathcal{D})\) be the parameters of the unlearned model produced by an unlearning algorithm \(\mathcal{U}\), which takes the original model \(\theta^o\), the forget set \(\mathcal{S}\), and the original dataset \(\mathcal{D}\), and outputs \(\theta^u\) via a post-processing mechanism.

\vspace{-3mm}
\subsection{General Machine Unlearning}

Building on this notation, the primary objective of a machine unlearning method is to eliminate the influence of a specified forget set \(\mathcal{S}\) from the trained model \(\theta^o\). 

\begin{definition}[(\(\varepsilon, \delta\))-unlearning \cite{triantafillou2024we}]
For a fixed dataset \(\mathcal{D}\), forget set \(\mathcal{S} \subseteq \mathcal{D}\), and a randomized
learning algorithm \(\mathcal{A}\), an unlearning algorithm \(\mathcal{U}\) is \((\varepsilon, \delta)\)-unlearning with respect to \((\mathcal{D}, \mathcal{S}, \mathcal{A})\) if for
all \(R \subseteq \mathcal{R}\), where \(\mathcal{R}\) denotes the output space (in this case, the space of model parameters \(\theta\)), we have:
\[
\Pr[\mathcal{A}(\mathcal{D} \setminus \mathcal{S}) \in R] \leq e^\varepsilon \Pr[\mathcal{U}(\mathcal{A}(\mathcal{D}), \mathcal{S}, \mathcal{D}) \in R] + \delta, \text{and}
\]
\[
\Pr[\mathcal{U}(\mathcal{A}(\mathcal{D}), \mathcal{S}, \mathcal{D}) \in R] \leq e^\varepsilon \Pr[\mathcal{A}(\mathcal{D} \setminus \mathcal{S}) \in R] + \delta.
\]
\end{definition}
The above definition expresses the degree of success of an unlearning algorithm $\mathcal{U}$ as a function of a notion of divergence between the distributions of the two models $\theta^r$ and $\theta^u$. 
Initially, a learning method $\mathcal{\theta^0}$ is trained on a dataset $\mathcal{D}$. Once a removal request or a set of requests is received, the unlearning process is initiated. The unlearning method $\mathcal{U}$ removes the impact of the forget set $\mathcal{S}$ and computes the updated model $\theta^r$ or $\theta^u$ accordingly. 
Based on the above definition, machine unlearning methods can be broadly classified into two categories: \emph{exact unlearning} and \emph{approximate unlearning}. The unlearning is said to be exact when \(\varepsilon = 0\) and \(\delta = 0\), meaning the output distributions of the unlearning algorithm and retraining are identical. In contrast, non-zero values of \(\varepsilon\) and \(\delta\) characterize approximate unlearning, with the degree of approximation increasing as these parameters grow. Notably, when \(\varepsilon\) and \(\delta\) are sufficiently small, the distributions of the retrained model \(\theta^r = \mathcal{A}(\mathcal{D} \setminus \mathcal{S})\) and the unlearned model \(\theta^u = \mathcal{U}(\mathcal{A}(\mathcal{D}), \mathcal{S}, \mathcal{D})\) become nearly indistinguishable:
\[
\lim_{\epsilon,\delta\rightarrow 0} Pr[\mathcal{U}(\mathcal{A}(\mathcal{D}), \mathcal{S}, \mathcal{D}) \in R] = \Pr[\mathcal{A}(\mathcal{D} \setminus \mathcal{S}) \in R]\;\forall R\in\mathcal{R},
\]
indicating a high-quality unlearning outcome.

The reason behind the above classification is the degree to which the unlearned model replicates the distribution of a retrained model. In \emph{exact unlearning}, the output of the unlearning algorithm is identically distributed to that of retraining from scratch on the pruned dataset, requiring $\varepsilon = 0$ and $\delta = 0$. This level of precision, while ideal, is often computationally infeasible in practice, e.g. large foundational models, and thus is usually reserved for privacy-critical applications. \emph{Approximate unlearning} relaxes this constraint by allowing small, bounded differences between the two distributions, captured by non-zero $\varepsilon$ and $\delta$. This formulation enables practical implementations while still providing quantifiable control over residual influence.

\begin{table*}[t]
    \centering
        \caption{Notation descriptions.}
        \vspace{-1ex}
    \begin{tabular}{|l|l|l|l|}
    \hline
        \textbf{Notation} & \textbf{Description} &\textbf{Notation}&\textbf{Description} \\ \hline
         $G$& A simple undirected graph &$\mathcal{G}$& A set of graphs \\ 
         $V$& A set of nodes in $G$& $E$& The set of edges in $G$ \\
         $v$& A node $v \in V$&$n,m$& The number of nodes, edges in $G$ \\
         $x$& A node feature vector &$y$ & The node label \\
         $\mathcal{Y}$&A set of node labels& $L$& A set of graph labels \\
         $X \in \mathbb{R}^{n \times d}$&The node feature matrix&$X_e \in \mathbb{R}^{m \times d^\prime}$& The edge feature matrix \\
         $d$& The size of node feature& $d^\prime$& The size of edge feature \\
         $\mathcal{D}$&An embedding dataset obtained from $\mathcal{G}$& $\mathcal{D}^\prime$& The dataset with removed data points \\
         $\theta^0$& Weights of a model trained on $\mathcal{D}$& $\theta^r, \theta^u$ & Weights of the \underline{r}etrained/\underline{u}nlearned models for $\mathcal{D}^\prime$ \\
         $\mathcal{A}$&A learning algorithm& $\mathcal{U}$& A machine unlearning method \\
         $h_v$& A node-level embedding& $h_G$& The graph-level embedding \\
         $\mathcal{N}_v$&Node $v$'s neighbors&$X_{\setminus v}$& The updated 
         $X$ after $v$'s removal \\
         $V^-,E^-$& A set of nodes/edges requested for removal&$H$& Hessian matrix \\ \hline
    \end{tabular}
    \label{tab:notations}
    \vspace{-3mm}
\end{table*}

\subsection{Graph Machine Learning}
We first introduce the relevant concepts on graph machine learning, which serves as background for the following sections.

\begin{definition}[Node Classification]
    Given a training graph $G$ and a set of labels $\mathcal{Y}$, the node classification task aims to learn a representation vector $h_v$ for each node $v \in V$ using a function $f$ such that $v$'s label can be predicted as $y_v = f(h_v)$, where $y_v \in \mathcal{Y}$ . 
\end{definition}

\begin{definition}[Link Prediction]
    Given a training graph $G$, the link prediction task aims to learn a representation vector $h_v$ for each node $v \in V$  using a function $\phi$, such that the existence of an edge between any two nodes can be predicted as $e_{u,v} = \phi(h_u, h_v)$. 
\end{definition}


\begin{definition}[Graph Classification]
    Given a set of graphs $\mathcal{G}$ and a set of labels $L$, the objective of graph classification task is to learn a representation vector $h_G$ for each $G \in \mathcal{G}$ using a function $g$ such that $G$'s label can be predicted as $y_G = g(h_G)$ where $y_G \in L$. 
\end{definition}

Unlike node classification and link prediction tasks, a set of graphs as represented by $\mathcal{G}$ is used to train the model for graph classification. This is due to the fact that here we aim to predict a label for the entire graph compared to a node.
The most common application of graph classification is molecular property prediction and toxicity identification \cite{hu2020open}. 

Similarly, graph regression and node regression problems share similarities with the graph classification and node classification tasks, but their primary focus lies in predicting continuous labels for each graph or node, rather than discrete classes. These machine learning tasks on graph-structured data have garnered significant attention and found diverse applications in recent years. With these definitions in mind, we now turn our attention to defining unlearning tasks on graphs.


\subsection{Graph Unlearning}

\begin{figure}[t]
    \centering
    \includegraphics[width = 0.4\textwidth]{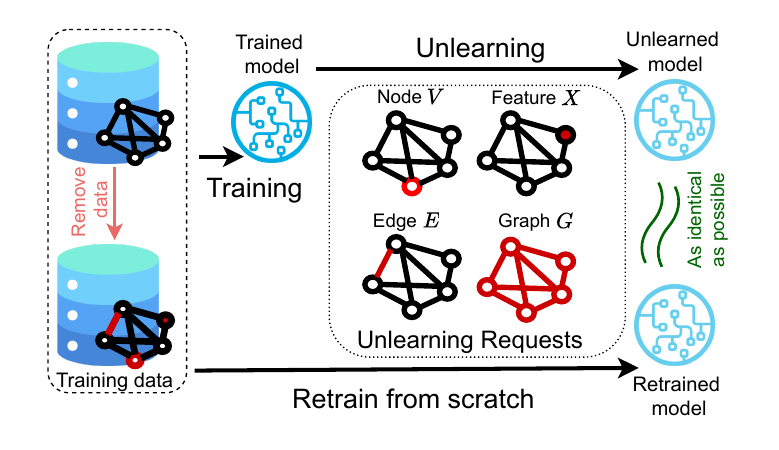}

    \vspace{-5mm}
    \caption{The graph unlearning framework is illustrated. Each removal request represents different scenarios where specific elements are to be removed from the trained model. Red color denotes the removal of component. }
    \label{fig:graph-unlearning}
    \vskip -2ex
\end{figure}

\begin{definition}[Graph Unlearning]
Graph unlearning is an essential extension of the unlearning process to the domain of GML, where the input is an attributed graph $G$. Formally, let \(\theta^0\) be the weight of the original graph machine learning model, trained on the graph \(G\) or a set of graphs $\mathcal{G}$ to perform a specific downstream task. The graph unlearning process aims to update $\theta^0$ in the presence of data removal in the form of node or node feature vector, or edge or edge feature vector, etc. An illustration of graph unlearning is shown in Figure \ref{fig:graph-unlearning}.
\end{definition}

\noindent{\textbf{Transductive Learning.}} 
Graph machine learning when training on/for a single training graph $G$. 

%
\begin{definition}[Transductive Node Unlearning]
Transductive node unlearning aims to effectively remove the data associated with an individual node, denoted as \(v\), from the trained model $\theta^0$ and its corresponding training graph \(G\). When a data subject \(v\) exercises the right to revoke their data, node unlearning ensures the erasure of \(v\)'s node features, represented by \(X_{\setminus v}\), as well as the links connecting \(v\) with other nodes, denoted by \(e_{u, v}\) for all \(u\) in the neighborhood \(\mathcal{N}_v\). Formally, node unlearning involves deriving an unlearned model \(\mathcal{M}^\prime\) that is trained on the updated graph \(G_{\setminus v} = G \setminus \{v, X_{\setminus v}, e_{u,v} | \forall u \in \mathcal{N}_v\}\). 
\end{definition}
\begin{definition}[Transductive Node Feature Unlearning]
Transductive node feature unlearning aims to obtain an unlearned model $\theta^r$ that is trained on the modified graph \(G_{\setminus v} = (V, E, X_{\setminus v}, W)\), where \(X_{\setminus v}\) represents the updated node feature matrix after removing the feature vector of node \(v\). The removal from $X$ can be done by replacing the feature vector with zeros or ones. The model $\theta^r$ is optimized to adapt to this updated graph \(G_{\setminus v}\) while ensuring that the influence of \(X_{\setminus v}\) on the model predictions is significantly reduced.
\end{definition}
\begin{definition}[Transductive Edge Unlearning]
Transductive edge unlearning refers to the removal of the influence of a specific edge from the trained model $\theta^0$ and its corresponding training graph \(G\), while retaining all nodes in the graph. In edge unlearning, a particular edge, denoted as \(e_{(u,v)}\) connecting nodes \(u\) and \(v\), is requested to be revoked by the data subject or individual to whom it pertains. This process involves removing the edge along with its corresponding attributes
from the model's training graph \(G\), without affecting other connections and node features. Formally, edge unlearning seeks to obtain an unlearned model $\theta^r$ that is trained on the updated graph \(G_{\setminus e_{(u,v)}} = (V, E \setminus \{e_{(u,v)}\}, X, X_{\setminus e_{(u,v)}})\), where
\(X_{\setminus e_{(u,v)}}\) corresponds to the updated edge feature matrix.
\label{def:transductive-edge}
\end{definition}

\textbf{Inductive Learning.}
Graph machine learning when seeking to generalize beyond a single graph to completely unseen nodes or even new graphs. The following types of unlearning requests can be received in this scenario:

%
%

\begin{definition}[Inductive Node Unlearning]
Inductive node unlearning responds to the query to remove the data associated with a set of nodes \(V^-=\{v_i,\dots,v_k\}\) from their corresponding training graphs \(\mathcal{G}'=\{G_i,\dots,G_k\}\) in the dataset. Similar to the transductive equivalent, inductive node unlearning ensures the erasure of \(v_i\)'s node features, and the links connecting \(v_i\) with adjacent nodes within graph $G_i$. Denoting $G'_{i}$ as the altered graph $G_i$ with node $v_i$ removed as described, we can restate the new version formally: inductive node unlearning involves deriving an unlearned model \(\mathcal{M}^\prime\) that is trained on the updated dataset \(\mathcal{G}_{\setminus V^-} = \mathcal{G} \setminus \mathcal{G}'\;\cup\;\{G'_{i},\dots G'_{k}\}\). 
\label{def:inductive-node}
\end{definition}

\begin{definition}[Inductive Edge Unlearning]
Straightforwardly similar to Definition~\ref{def:inductive-node}, inductive edge unlearning responds to the query to remove the data associated with a set of edges \(E^-=\{e_{(u_i,v_i)},\dots,e_{(u_k,v_k)}\}\) from their corresponding training graphs \(\mathcal{G}'=\{G_i,\dots,G_k\}\) in the dataset. Let $G'_i$ be the altered graph $G_i$ with edge $e_{(u_i, v_i)}$ removed as specified in Definition~\ref{def:transductive-edge}, inductive edge unlearning involves deriving an unlearned model \(\mathcal{M}^\prime\) that is trained on the updated dataset \(\mathcal{G}_{\setminus E^-} = \mathcal{G} \setminus \mathcal{G}'\;\cup\;\{G'_{i},\dots G'_{k}\}\). 
\end{definition}

\begin{definition}[Graph-Level Unlearning]
This is a special case to the inductive setting, where the training dataset has more than one graph. A request to unlearn a set of graphs $\mathcal{G}^-=\{G_i^-,\dots,G_k^-\}$ requires updating the model parameters to account for the removal of every graph in \(\mathcal{G}^-\) and its associated information from the set of graphs \(\mathcal{G}\). Formally, graph-level unlearning involves deriving an unlearned model \(\mathcal{M}^\prime\) that is trained on the updated dataset \(\mathcal{G}_{\setminus G^-} = \mathcal{G} \setminus \mathcal{G}^-\). 
\end{definition}

%
%
%

%
%
When dealing with graph-level unlearning requests, the task becomes similar to conventional machine unlearning, as each graph is treated as an independent data point in such scenarios. Consequently, well-established machine unlearning approaches can be effectively applied in these cases \cite{pan2023unlearning}. However, it is important to note that this approach has certain limitations when it comes to unlearning graphs, as it may restrict the possibility of accommodating other types of unlearning request simultaneously. For instance, node and edge unlearning requests can be received in the graph-level unlearning tasks which necessitates dedicated efforts beyond the simple unlearning. Moreover, unlearning in the domain of graphs introduces a unique and intricate challenge, setting it apart from traditional machine unlearning in other domains. The fundamental distinction lies in the underlying message passing mechanism, where information is iteratively propagated through multiple hops across the graph's interconnected nodes. Once a node or edge is removed from the graph, unlearning extends beyond simple data point removal, as it necessitates updating the effects of the deleted nodes throughout their respective neighborhoods. This complexity arises from the convolution operation inherent in graph representation learning, where information spreads across the graph's structure. Consequently, the unlearning process demands careful handling of the model's parameters to ensure the accurate recalibration of neighboring nodes and edges after data removal.

\vspace{-2mm}
\section{Categorization and Frameworks}
\label{sec:categorization}
%
%
Drawing from the existing literature, we classify graph unlearning into two distinct categories: exact unlearning, and approximate unlearning. In Table \ref{tab:taxanomy}, we offer a concise overview of these methods, providing a glimpse into their essence. In the forthcoming paragraphs, we will delve into each of these categories, providing explanations of the methodologies employed in graph unlearning. Our aim is to shed light on the intricacies of these methods, uncovering their unique contributions to the fascinating domain of graph unlearning.

\begin{table*}[!t]
\tiny
 \caption{\small Categorization of graph unlearning approaches. $V$ denotes node unlearning, $E$ represents edge unlearning, $X$ refers to node feature unlearning, $G$ represents the graph unlearning approach.
 Approx. unl. stands for approximate unlearning, 
\textbf{Code} indicates the availability of the implementation for the proposed method.
 } 
\begin{tabular}{|l|c|c|c|c|c|c|c|c|p{8.2cm}|}
\hline
Method                                  & \textbf{Year \& Venue}&\textbf{Code}&$\boldsymbol{V}$ & $\boldsymbol{E}$ & $\boldsymbol{X}$ & $\boldsymbol{G}$ & \textbf{ $G$ Type} & \textbf{ $\mathcal{M}$ Type} & \textbf{Method Summary} \\ \hline 
GraphEditor~\cite{cong2022grapheditor}  &arXiv 2022& \redcross &  \greentick   & \greentick     &  \greentick       &   \redcross    &     undirected   & exact unl.  &  Compute closed-from solution of GNN output and $\mathcal{Y}$ and update the model for non-convex setting              \\ \hline
  RecEraser~\cite{chen2022receraser} & WWW 2022 & \greentick & \greentick &\greentick &\redcross &\redcross & bipartite & exact unl.& SISA-based framework to preserve the collaborative information in data partition\\ \hline
Untrain-ALS~\cite{xu2023netflix} & arXiv 2023 &\redcross &\greentick &\greentick & \redcross& \redcross& bipartite & exact unl. & Retrain with Alternating Least Squares  \\ \hline
UltraRE~\cite{li2024ultrare} & NeurIPS 2024 & \greentick & \greentick& \greentick& \redcross& \redcross& bipartite &exact unl. & SISA-based framework with balanced data partition and simplified combination process\\ \hline
GraphRevoker~\cite{zhang2024graphunlearningefficientpartial} & WWW 2024 & \redcross & \greentick & \greentick & \greentick & \redcross & undirected & exact unl. & SISA-based method with graph-aware clustering and contrastive submodel aggregation. \\ \hline

SCIF~\cite{li2023selective} & ESWA 2023 &\redcross &\greentick &\greentick& \redcross& \redcross& bipartite & approx. unl.& Selective and collaborative influence function in recommendation\\ \hline
RRL~\cite{you2024rrl} & AAAI 2024& \redcross &\greentick &\greentick& \redcross& \redcross& bipartite & approx. unl.& Apply reverse learning in recommendation\\ \hline
IFRU~\cite{zhang2023recommendation} & ACM ToRS 2024 &\redcross & \greentick &\greentick& \redcross& \redcross& bipartite & approx. unl.& Influence function-based recommendation unlearing\\ \hline

GraphEraser~\cite{chen2022graph} & SIGSAC 2022  &\greentick&\greentick   & \greentick     &  \redcross       &   \redcross    &     undirected       &  approx. unl. &Cluster the graph, train a separate model on each cluster, and then aggregates the results            \\ \hline
CGU~\cite{chien2022certified} & NeurIPS 2022 & \greentick&\greentick    &\greentick      &  \greentick       &  \redcross     &  undirected &    approx. unl.     &  Use Hessian for updating the model and provides theoretical foundation              \\ \hline
GUIDE~\cite{wang2023inductive}  & USENIX 2023 &\greentick&\greentick    &  \redcross    &     \redcross    &    \redcross   &    undirected  &   approx. unl.   &  Compute $k$ shards, apply repair function to recover edges, train $k$ models and aggregate               \\ \hline
CEU~\cite{10.1145/3580305.3599271} & KDD 2023 & \greentick & \redcross &  \greentick    & \redcross        & \redcross      &   undirected         &  approx. unl.  & Propose a custom edge influence function and Newton-update accordingly           \\ \hline

Unlearning GST~\cite{pan2023unlearning}  & WWW 2023 &\greentick&\redcross    &  \redcross    &  \redcross       &   \greentick    &     undirected  &  approx. unl.   &  Use the traditional influence functions to compute the change and update the model              \\ \hline
GIF~\cite{wu2023gif}  & WWW 2023&\greentick&\greentick     &   \greentick   &  \greentick       &  \redcross     &     undirected      & approx. unl.&  Use the traditional influence functions and incorporate addition term to the loss               \\ \hline
Projector~\cite{cong2023efficiently} &AISTATS 2023 &\greentick&  \redcross    &   \redcross   & \greentick        &  \redcross     &    undirected &    approx. unl.   &   Use orthogonal projection as a weighted combination of node features for unlearning             \\ \hline
GNNDelete~\cite{cheng2023gnndelete}  & ICLR 2023 & \greentick&\greentick   &\greentick      & \greentick        & \redcross      &     undirected &   approx. unl.    &  Introduce shared weight matrices across nodes; apply layer-wise deletion operator to update the model              \\ \hline
SGC~\cite{chien2022efficient}  & ICLR 2023 & \greentick&\greentick   &\greentick      & \greentick        & \redcross      &     undirected &   approx. unl.    &  Use influence function for model updates, derive robust theoretical guarantees in convex setting              \\ \hline
 SAFE~\cite{dukler2023safe} & ICCV 2023 & \redcross & \greentick&\greentick&\redcross&\redcross&undirected&approx. unl.& Using graph sharding mechanism to train secure GNN models \\ \hline

GUKD~\cite{DBLP:conf/icics/ZhengLWL23} & ICICS 2023 & \redcross & \greentick & \redcross & \redcross & \redcross & undirected & approx. unl. & Finetune the unlearning model with soft labels from a smaller independently retrained model. \\ \hline
MEGU~\cite{li2024effectivegeneralgraphunlearning} & AAAI 2024 & \greentick & \greentick & \greentick & \greentick & \redcross & undirected & approx. unl. & Simultaneously unlearn and preserve original utility through a set of high influence nodes
\\ \hline
D2DGN~\cite{sinha2024distilldeleteunlearninggraph} & arXiv 2024 & \redcross & \greentick & \greentick & \greentick & \redcross & undirected & approx. unl. & Distill model to destroy knowledge of the unlearned data and retain knowledge of the remaining data \\ \hline
IDEA~\cite{10.1145/3637528.3671744} & KDD 2024 & \greentick & \greentick & \greentick & \greentick & \redcross & undirected & approx. unl. & Provides a certification for edge, node, and feature unlearning using influence function \\ \hline
UtU~\cite{tan2024unlink} & WWW 2024 & \greentick & \greentick & \redcross & \redcross & \redcross & undirected & approx. unl. & Prevent message passing by removing unlearned edges during inference \\ \hline
Cognac~\cite{kolipaka2025cognacshotforgetbad} & ICML 2025 & \greentick & \greentick & \greentick & \redcross & \redcross & undirected & approx. unl. & Detect nodes affected by the unlearning requests and apply contrastive unlearning on them \\ \hline
ETR~\cite{yang2024erase} & AAAI 2025 & \greentick & \greentick & \greentick & \greentick & \redcross & undirected & approx. unl. & Edit model weights using Fisher information matrix, then one-step gradient update for remaining data \\ \hline
ScaleGUN~\cite{yi2025scalable} & ICLR 2025 & \greentick & \greentick & \greentick & \greentick & \redcross & undirected & approx. unl. & Maintain certification for large graphs by lazy local propagation and only retrain when exceeding tolerance threshold \\ \hline
KGU Schema~\cite{xiao-etal-2025-knowledge} & COLING 2025 & \greentick & \greentick & \greentick & \redcross & \redcross & undirected & approx. unl. & Query subgraphs similar to the removal request and unlearn them \\ \hline
AGU~\cite{ding2025adaptivegraphunlearning} & IJCAI 2025 & \redcross & \greentick & \greentick & \greentick & \redcross & undirected & approx. unl. & Address edge removal and feature removal separately by identifying and correcting affected nodes \\ \hline
SGU~\cite{li2025scalablegraphunlearningnode} & arXiv 2025 & \redcross & \greentick & \greentick & \greentick & \redcross & undirected & approx. unl. & Scalable GU through a set of high influence nodes with Node Influence Maximization \\ \hline
\end{tabular}
\label{tab:taxanomy}
\vspace{-1mm}
\end{table*}

\vspace{-2mm}
\subsection{Exact Graph Unlearning}

Exact graph unlearning seeks to achieve complete unlearning (i.e., ensure that removal data is fully unlearned) by retraining from scratch. However, naive retraining can lead to significant computational costs. To enhance efficiency, research has focused on the following two key directions.

\subsubsection{SISA-based Retraining}

A representative exact unlearning algorithm is an ensemble framework called SISA (Sharded, Isolated, Sliced, and Aggregated) \cite{bourtoule2021machine}. Unlike traditional retraining which relies on the entire training dataset, SISA randomly partitions the training dataset into distinct and disjoint shards, each independently trained to form a set of shard models. Once a removal operation happens, only the relevant shard models need to be retrained. 
In recommendation scenarios, the SISA framework has been tailored to handle the user-item bipartite graph \cite{chen2022receraser}. To address challenges in data partitioning, RecEraser introduces three novel partitioning algorithms that preserve collaborative information, which is crucial for recommendation. Specifically, these three partitions are based on the similarities of users, items, and interactions. Additionally, the partition strategies are designed to achieve a balanced partition for the unlearning efficiency. Following RecEraser, \cite{li2024ultrare} rethinks from an ensemble-based perspective and further refines RecEraser from the three potential losses respectively (i.e., redundancy, relevance, and combination). In detail, UltraRE \cite{li2024ultrare} integrates transport weights in the clustering algorithm to optimize the data partition process, and simplifies the combination estimator with simple architecture. GraphRevoker~\cite{zhang2024graphunlearningefficientpartial} took a different approach to prevent information loss by proposing a different clustering algorithm named graph-aware sharding, and improving upon submodel ensembling with contrastive submodel aggregation.

However, note that some SISA-style methods contain modifications that made them no longer exact. Specifically, if the graph partitioning process is not fully random and instead clustered conditionally on edge or feature information, it voids the exactness property of the overall method.
One example is GraphEraser \cite{chen2022graph}
which begins by partitioning the original training graph into separate and disjoint shards. Each shard is then used to train an individual model. When a node requires a prediction, GraphEraser forwards to all shard models and obtains their corresponding posteriors. These posteriors are then aggregated each shard model, enabling GraphEraser to make a robust prediction. Furthermore, when a node makes an unlearning request, GraphEraser promptly removes the node from the corresponding shard and subsequently retrains the shard model. This process ensures that the influence of this node is mitigated, and the model remains updated and accurate without requiring a complete retraining of the entire graph
machine learning model. The difference to SISA mainly lies in the data partition and aggregation components. GraphEraser proposes two strategies to achieve balanced graph partition. One strategy is based on community detection, which aims to divide the graph into groups of nodes with dense connections internally and sparse connections between groups. Another strategy relies on embedding clustering, which considers both the node feature and graph structural information for the partitioning. Nodes with similar embeddings are assigned to the same shard. To further ensure a balanced partition, a Balanced Embedding k-means algorithm is devised. Additionally, considering that the shard models contribute differently to the final prediction, GraphEraser propose a learning-based aggregation method that optimizes the importance score of each shard model, which reflects its significance in the overall prediction process.

Inspired by GraphEraser, Wang et al. present a method named GUIDE (GUided InDuctivE Graph Unlearning)~\cite{wang2023inductive}. GUIDE consists of three essential components: guided graph partitioning, graph repairing, and aggregation. GUIDE begins by partitioning the graph into shards using newly proposed clustering techniques. Subsequently, GUIDE employs subgraph repair methods to restore lost connections and integrates them into their respective shards. This step is crucial to reconstruct the graph after partitioning where some of the connections are lost. Finally, GUIDE applies an aggregation method to learn a similarity score for each shard. This similarity score is then utilized for making inferences during the unlearning process.

\subsubsection{Parameter-based Retraining}

Another way to implement exact unlearning is to store the historical parameters of the model for further adjustment.
\cite{ullah2021machine} propose an unlearning approach for training the model using mini-batch stochastic gradient descent (SGD). Throughout the training process, the model parameters are saved at each iteration. When deletion requests are received, retraining commences solely from the iteration when the deleted data first appeared. This strategy optimizes the unlearning process, allowing for targeted updates and minimizing redundant retraining efforts. 
The authors in \cite{cong2022grapheditor} introduce GraphEditor, a graph representation learning and unlearning approach that supports both node and edge deletion and addition. GraphEditor offers an exact unlearning solution without necessitating extensive retraining. The core principle behind GraphEditor lies in transforming the conventional GNN training problem into an alternative problem with a closed-form solution. Once the initial training is completed and the weight matrix \(W\) is obtained, GraphEditor computes the closed-form solution, returning \(W^\prime\), the updated weight matrix, and \(S^*\), the inverse correlation matrix for unlearning. In the event of a deletion request, GraphEditor updates the model parameters \(W^\prime\) by first removing the effect of the requested node/edge on the remaining nodes and then updating the model. This targeted approach ensures efficient and precise unlearning, making GraphEditor a useful tool for managing graph data and maintaining model integrity. However, a key limitation of the paper lies in its applicability, as the majority of GNNs employed are non-linear. Consequently, the current results may have limited relevance in such scenarios. Furthermore, envisioning the generalization of the current approach to more intricate network structures presents challenges, as it heavily relies on the linear characteristics of the problem. 
In recommendation scenario, Untraining Alternating Least Squares (Untrain-ALS) \cite{xu2023netflix} proposes to fine-tune based on the pretrained weights. They retrain the model with new data after the removal until convergence. They show that mathematically retraining with ALS equals minimizing the model loss on the remaining data after removal, making it an exact unlearning method.
\vspace{-3mm}
\subsection{Approximate Graph Unlearning}

Approximate unlearning usually aims to estimate the influence of unlearning data, and directly removes the influence through parameter manipulation. While facing the challenges related to unlearning completeness, these methods are generally more efficient. We discuss them in the convex and non-convex settings in the following.

\subsubsection{Convex Setting}
A common assumption for theoretical guarantees of unlearning method is (strong) convexity, where the Hessian of the loss function with respect to the input is positive (semi-)definite. Rigorously, for loss function $L$, model weights $W_1, W_2$, and a dataset $\mathcal{D}$, convexity is defined as:
\begin{equation*}
    L(\lambda W_1 + (1-\lambda) W_2, \mathcal{D}) \le L(\lambda W_1, \mathcal{D}) + L((1-\lambda) W_2, \mathcal{D}),
\end{equation*}
for all $0\le\lambda\le 1$. For \textit{strong} convexity, the weak inequality above becomes a strict less-than ($<$).
When the convexity assumption holds, the loss landscape of the model contains exactly one unique minimum, which is also a global minimum---making proving mathematical bounds vastly simpler. However, for this powerful condition, we trade off applicability: convexity condition is very strict, and mostly only apply on linear model with simple loss functions.


 The paper \cite{gama2018diffusion} introduces a graph embedding technique called Graph Scattering Transform (GST). GST relies on a collection of multiresolution graph wavelets, a pointwise nonlinear activation function, and a low-pass operator. Building upon GST, \cite{pan2023unlearning} propose an interesting graph unlearning approach, aptly named Unlearning GST. By utilizing GST embeddings, Unlearning GST defines an unlearning model denoted as $\mathcal{M}^\prime$, which updates the trained model from $W$ to $W^\prime$. The latter represents an approximation of the unique optimizer of $L(W, \mathcal{D}^\prime)$. The derivation of $W^\prime$ is as follows:

To begin, $H_{W}$ is defined as the Hessian of $L(., \mathcal{D}^\prime)$ at $W$ and is computed as:
$$H_{W} = \nabla^2L(W, \mathcal{D}^\prime)$$

Next, the gradient difference is evaluated by computing the difference between $\nabla^2L(W, \mathcal{D})$ and $\nabla^2L(W, \mathcal{D}^\prime)$:
\begin{equation}
\label{eq:gradientupdate}
\bigtriangleup = \nabla^2L(W, \mathcal{D}) - \nabla^2L(W, \mathcal{D}^\prime)    
\end{equation}

Using the obtained gradient difference, $W^\prime$ is derived as:
$$W^\prime = W + H_{W} \bigtriangleup$$

$W^\prime$ thus represents the updated model obtained through this unlearning process. However, this approach has a limitation related to the computation of the wavelet transform, which is cubic for node removal and quadratic for node feature removal.

A notable addition to the research landscape, akin to the aforementioned \cite{gama2018diffusion} study, is the certified graph unlearning approach presented in \cite{chien2022certified}. Inspired by the principles laid out in \cite{guo2020certified}, this study extends the unlearning mechanism to accommodate graphs within a linear GNN setting. Generalizing the bounds, the authors obtain certified unlearning guarantees for node features, nodes, and edges. Despite its significance, this work's main limitation lies in its focus on linearity in GNNs, which limits its applicability to a select set of applications.

In the graph-level unlearning context, each graph effectively becomes an independent data point, and the task of unlearning translates to removing or updating individual graphs from the model. This interesting transformation aligns the graph-level unlearning problem with the established principles of machine unlearning. As a result, the conventional methodologies and techniques developed for machine unlearning hold relevance and can be effectively applied to tackle the graph-level unlearning problem \cite{pan2023unlearning}.

Building upon the foundation laid by the certified data removal approach in \cite{guo2020certified}, an approximate unlearning approach has been proposed in \cite{chien2022efficient}. This study introduces methods for unlearning node features, edges, and nodes, presenting robust theoretical guarantees within the context of limited GNN models operating in a convex setting. The crux of their work lies in the utilization of influence function, combined with the incorporation of graph information through an additional term in the gradient update function. Moreover, the paper establishes robust bounds on the gradient residual norms, elevating the efficacy of graph unlearning in three distinct scenarios.

The authors in \cite{wu2023gif} introduce another unlearning method, Graph Influence Function (GIF). GIF aims to model the influence of each training data point on the model with respect to various performance criteria and subsequently eliminates the negative impact. The method incorporates an additional loss term that considers the influence of neighbors and estimates parameter changes in response to a $\epsilon-$mass perturbation in deleted data. Note that while GIF can practically be applicable to the non-convex setting, the authors highlighted the decrease in its performance on sufficiently deep models as the setting deviates far enough from the theoretical assumptions. Another similar approach is ``Projector'' by Cong et al. \cite{cong2023efficiently} This method aims to remove the impact of deleted data from the learned model by projecting the weight parameters to a different subspace that is unrelated to the deleted data. Specifically, Projector is designed to work with linear GNNs, and demonstrates that all gradients lie within the linear span of all node features. Leveraging this insight, it performs unlearning using an orthogonal projection represented as a weighted combination of the remaining node features. By applying this technique, the learned model can effectively adapt to the removal request and update itself while minimizing the impact of the deleted data.
IDEA~\cite{10.1145/3637528.3671744} generalizes certified unlearning to all types of unlearning requests. Cognac~\cite{kolipaka2025cognacshotforgetbad} brings the contrastive approach to unlearning and applies them on nodes that are affected by the removal request, but requires the graph to be homophily and the model to be homophily-preserving.

\subsubsection{Non-Convex Setting}

In the pursuit of approximate unlearning within the non-convex setting, CEU is introduced in \cite{10.1145/3580305.3599271}. CEU specifically targets edge unlearning in graphs. To achieve this, CEU focuses on reversing the influence of the edge that needs to be removed from the new model $W^\prime$, leveraging the concept of influence functions \cite{pmlr-v70-koh17a}. The authors propose an estimation of the influence function by upweighting the set of all affected nodes and subsequently calculate the reverse of the Hessian matrix multiplied by the gradient vector to obtain the influence. To address computational complexity, they employ conjugate gradient optimization, effectively reducing the computational burden. 
ETR~\cite{yang2024erase} follows a similar approach, but edit weights accordingly to the Fisher information matrix, which is computed approximately and efficiently.

The unlearning process, as explored in \cite{mitchell2022fast}, can potentially harm the performance of the underlying predictive model. In an effort to mitigate this issue and avoid updating the trained model, \cite{cheng2023gnndelete} introduces GNNDelete. This approach formalizes two essential properties for the GNN deletion method: \textit{deleted edge consistency} and \textit{neighborhood influence}. Deleted edge consistency ensures that the predicted probabilities for deleted edges in the unlearned model stay similar to those for nonexistent edges. Similarly, neighborhood influence guarantees that predictions in the local vicinity of the deletion retain their original performance and remain unaffected by the removal. In pursuit of efficiency and scalability, GNNDelete employs~a layerwise deletion operator to modify a pretrained GNN model. Upon receiving deletion requests, GNNDelete freezes the existing model weights and introduces small, shared weight matrices across the nodes in the graph. GNNDelete guarantees strong performance by ensuring the difference between node representations obtained from the trained model $\mathcal{M}$ and those revised by GNNDelete $\mathcal{M}^\prime$ remains theoretically bounded. 

Some approaches utilize knowledge distillation (KD) to unlearn specific subsets of the training data. Distill-to-Delete (D2DGN)~\cite{sinha2024distilldeleteunlearninggraph} takes inspiration from KD and distill the full model to a new unlearned one, using loss terms to either destroy or retain knowledge of some training data. Specifically, D2DGN uses the cross-entropy, then maximizes it for the unlearning data and minimizes for the rest. GUKD~\cite{DBLP:conf/icics/ZhengLWL23} flips the traditional KD settings and instead applies distillation to the unlearning model with soft label from an independently retrained model on the remaining data. This retrained teacher model belongs to a smaller architecture to reduce computational overhead.

Instead of finding the most affected weights, which is the case for influence function-based methods, some work aim for detecting high influence nodes instead. MEGU~\cite{li2024effectivegeneralgraphunlearning} supports all type of removal requests by addressing the delete requests' effects on these important nodes. This is done through simultaneously optimizing two loss terms, where one unlearns the requested-for-deletion nodes, and another preserves original utility of the model, on the important nodes only. AGU~\cite{ding2025adaptivegraphunlearning} separately addresses edge removal and feature removal, and in turn supports multiple types of graph unlearning by some combination of the two above. SGU~\cite{li2025scalablegraphunlearningnode} further improves on unlearning capability and performance by proposing a better node selection method named Node Influence Maximization, which is much faster than other competing baselines.

Newer works focus more on substantially reducing runtime for graph unlearning by simply not doing computation. Unlink-to-Unlearn (UtU)~\cite{tan2024unlink} prevents message passing of the unlearning nodes by removing unlearned edges during inference. UtU is very fast, and the authors conducted a thorough experiment to empirically verify its unlearning power. A more recent and notable method is ScaleGUN~\cite{li2025scalablegraphunlearningnode}, which works on all graph unlearning request types and deal with the certified unlearning setting. It accomplishes this by determining loss bounds that would break the unlearning certification, and only commits lazy local propagation until when that bound is crossed.

Regarding specialized approximate unlearning for specific applications, in the knowledge graph setting,
{KGU Schema~\cite{xiao-etal-2025-knowledge} queries subgraphs similar to the removal request based on the graph entity schema, and only commit unlearning on them.}
For the recommendation setting, RRL \cite{you2024rrl} is the initial work to apply reverse learning in recommendation. Based on the original recommendation objective, they devise a reverse objective. They train this newly proposed reverse objective to remove the impact of the target data. Selective and Collaborative Influence Function (SCIF) \cite{li2023selective} is applied in recommendation to perform reverse gradient operations on the learned model. It directly removes the influence of target data by reverse operation, avoiding any kind of retraining. To further reduce the computational cost, it select the most pertinent user embedding for the reverse operation. Additionally, to enhance model utility, the collaborative data of the target data is considered as the collaboration preservation. This work quantifies the direct influence of unusable data on the optimization loss. However, the unusable data also affect the computational graph of recommender models such as LightGCN. This inspires a new Influence Function-based Recommendation Unlearning (IFRU) framework \cite{zhang2023recommendation}, focusing on the extension of the influence function to quantify the influence of unusable data in the computational graph aspect.


\section{Graph Unlearning Evaluation}

In this section, we summarize the evaluation of graph unlearning, focusing on three key aspects: \textit{unlearning completeness}, \textit{unlearning efficiency}, and \textit{model utility}. Additionally, we list other less commonly measured aspects for a more comprehensive evaluation.

\subsection{Unlearning Completeness}
Unlearning completeness is the primary objective of the unlearning task, measuring whether the information from the target data has been fully removed. This concept is sometimes referred to as unlearning efficacy \cite{wu2023gif}. For exact unlearning approaches, the model is obtained through retraining after removing the data. Therefore, exact unlearning methods guarantee complete unlearning. For approximate unlearning methods, there is no such guarantee and proofs are required to demonstrate whether the model fully unlearn the target data.

\textbf{Membership Inference-based:} One direction is to measure completeness based on the performance of a membership inference attack (MIA). Membership inference is an attack that aims to determine whether a specific data point was part of the training dataset used to train a machine learning model. This becomes particularly relevant in machine unlearning and the success of MIA can serve as a suitable measure for the quality of unlearning \cite{cheng2023gnndelete,yeom2018privacy}. To evaluate, one can conduct the membership inference for the removed data. Ideally, the inference should show a high score for the original model, indicating the existence of these data in the training dataset. On the other hand, a low value should be presented for the unlearned model showing that these data is not used for training.

\textbf{Prediction Difference-based:} A complete unlearning algorithm should generate similar predictions to the retrained model. According to this criterion, completeness can be measured based on the difference of model output space between the retrained model and unlearned model. Let $\mathbf{W}^u$ and $\mathbf{W}^r$ be the model parameters of unlearning model and retrained model (these notations are also used in parameter difference-based evaluation below), the general difference can be measured by $d(f(\mathbf{W}^u), f(\mathbf{W}^r))$ where $f(\cdot)$ represents the model output (e.g., the final activation output) and $d(\cdot, \cdot)$ is a distance function. 
\cite{wu2022fast} measures the distance as the Jensen–Shannon divergence (JSD) between the posterior distributions output by these two models where a smaller JSD indicates a higher similarity between the two models in terms of their outputs. \cite{cong2022grapheditor} measures the Euclidean distance of final activation prediction on both deleted nodes and test nodes. Specifically, for nodes in deleted set and test set $\mathcal{B} \in \{\mathcal{V}_{\text{rm}}, \mathcal{V}_{\text{test}}\}$, they compare the distance of final activation as $\mathbb{E}_{v_i \in \mathcal{B}} \left[ \left\| \text{softmax}(\mathbf{x}_i \mathbf{W}^u) - \text{softmax}(\mathbf{x}_i \mathbf{W}^r) \right\|_2 \right]$. Similarly, \cite{cong2023efficiently} compare the distance between the final activation. In addition to the deleted set and test set, they also compare the performance in remaining node set $\mathcal{V}_{\text{remain}} = \mathcal{V}_{\text{train}} \setminus \mathcal{V}_{\text{rm}} $.

\textbf{Parameter Difference-based:} A complete unlearning algorithm should have similar parameter weights with the retrained model. The difference can be measured as $d(\mathbf{W}^u, \mathbf{W}^r)$.
Based on this assumption, \cite{cong2022grapheditor} measures the Euclidean distance between the parameters returned by the unlearning algorithms and the parameters obtained via retraining from scratch. \cite{cong2023efficiently} measures the difference between normalized weight parameters.

\textbf{Others:} Several 
other less commonly used
evaluation methods have been proposed. \cite{cong2022grapheditor} intentionally modifies the target node's label to an extra category and investigates whether the unlearned model classifies the node to this extra label. An extra binary feature is injected into the node features to help the model associate the deleted nodes with the extra-label category. Intuitively, a complete unlearning method should never predict the extra label. Similarly, \cite{cong2023efficiently} append an extra binary feature to all nodes and set the extra binary feature as 1/0 for the deleted nodes/other nodes respectively. They also add an extra category and change all deleted nodes to this extra category. They compare the weight norm of the injected channel before and after unlearning to determine whether the unlearning is complete.
\cite{wu2023gif} propose an evaluation method based on adversarial attack. They add a certain number of edges to the training graph where each added edge links two nodes from different classes. The adversarial noise is expected to reduce the performance due to misleading the representation learning. Thereafter, they use the adversarial edges as unlearning requests and evaluate the utility of the estimated model. A larger utility gain indicated a higher unlearning efficacy. \cite{zhu2023heterogeneous} uses the utility performance on the deleted data as a measurement for unlearning completeness, assuming a smaller utility performance on the forgetting sets indicates that the model forgets more thoroughly.

There remains ongoing debate about how to best evaluate unlearning completeness, arguing against the commonly-used measurements.
For instance, MIA-based methods' reliability depend on the quality of the underlying attack, which often fails for well-generalized models~\cite{wu2021modelextractionattacksgraph,rezaei2021difficultymembershipinferenceattacks}, providing a false sense of security. Moreover, as MIAs are logit-based, a low attack success rate may not imply that the underlying latent space is unlearned, and only the shallow classification layers are modified instead~\cite{kim2025trulyforgettingcriticalreexamination}. Prediction Difference and Parameter Difference methods both have a very simple yet practical failure mode, that is when there are duplicates or dependency in the data~\cite{ye2025dataduplicationnovelmultipurpose}. Parameter Difference may prove to be unreliable, as functionally-equivalent models can have very different weights~\cite{alon2025doesoverparameterizationaffectmachine}.
Some argue that unlearning is only well-defined at the algorithmic level, where the only possible auditable claim to unlearning is that they used a particular algorithm designed to allow for external scrutiny during an audit \cite{thudi2022necessity}.
\subsection{Unlearning Efficiency}
Retraining from scratch is considered the gold standard in unlearning, but in practice, it incurs a prohibitive computational overhead. In real-world applications, models need to adapt quickly to the removal request from users. Therefore, the unlearning efficiency becomes a crucial aspect to evaluate. Unlearning efficiency specifically refers to how rapidly a machine unlearning model can modify or remove specific data points or knowledge without compromising overall performance. Models are typically assessed by comparing the empirical time required to complete the unlearning process. In addition to the empirical time, the time complexity sometimes is calculated in algorithm analysis to present a computational complexity~\cite{chen2022graph}.

Overall, most methods' complexity scales linearly with the number of layers, and quadratic with the feature dimension~\cite{fan2025opengu}.

\subsection{Model Utility}
Graph unlearning is inherently a multi-objective task. In addition to erasing the target data, the unlearning method needs to prevent performance degradation on the remaining data~\cite{cong2022grapheditor}. Therefore, preserving model utility for downstream graph tasks is also a crucial objective.
Model utility refers to the assessment of how effective and reliable an unlearning model is in achieving its intended purpose. To evaluate this aspect, common utility metrics such as F1 score \cite{chen2022graph}, accuracy \cite{wu2022fast} for classification task, AUROC for link prediction \cite{cheng2023gnndelete}, Normalized Discounted Cumulative Gain (NDCG) and Hit Ratio (HR) for recommendation task \cite{chen2024cure4rec} are often employed. The specific utility metrics used depend on the nature of the downstream task. Model utility is frequently compared against the `scratch method' (i.e., training from scratch), which, although less efficient, provides a baseline for utility performance evaluation.

\subsection{Other Aspects}
Space efficiency is considered in \cite{cheng2023gnndelete} where the number of training parameters is measured and whether the training parameters will scale with respect to the graph size is discussed. Fairness aspect is discussed the first time in recommendation unlearning \cite{chen2024cure4rec}. Fairness is evaluated from two aspects including the fairness between active and inactive user groups, and the fairness among different shards. Both fairness metrics are calculated based on the average utility performance.

\section{Graph Unlearning Applications}
\label{sec:applications}

Graph unlearning stands as an important topic in the realm of machine learning and data privacy. Due to the exponential development of graph-structured data across multiple domains, ensuring the right to be forgotten and protecting individual privacy have become of the utmost importance. As we observe the remarkable success of graph machine learning applications, it is important that we devise effective graph unlearning strategies. By enabling the removal of all forgotten data while preserving the integrity of interconnected networks, graph unlearning gives individuals greater control over their data and nurtures confidence in the responsible management of data. 
In the next sections, we first present potential unlearning requests from a user's perspective in practice and then discuss the applications of graph unlearning in a variety of domains.

\subsection{Realworld Graph Unlearning from an Individual Perspective}

\begin{figure}
    \centering
    \includegraphics[width=0.48\textwidth]{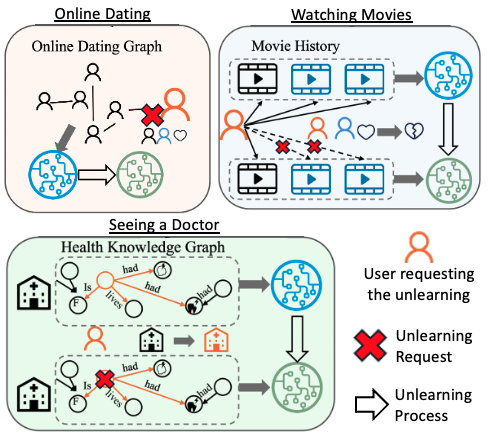}
    \caption{Real-world graph unlearning applications.}
    \label{fig:applications}
    \vspace{-2mm}
\end{figure}

Unlearning requests are increasingly common in practice. Figure \ref{fig:applications} shows three examples for a single individual. In the first, the individual uses an online dating application to seek a relationship, matches with a partner, and then deletes their account. In the second, after the breakup, they request removal of movie records—primarily chosen by the partner—that they had watched together, so the platform stops recommending similar titles. In the third, the individual transfers to a new hospital and asks that their previous health records be removed from the original facility’s models. In each case, the request originates from the user, and the trained model must unlearn the specified information to comply.

\subsection{Social Networks}

The application of graph unlearning in social networks is essential for mitigating the risk that a trained model may inadvertently learn sensitive information about users' social relationships. On the basis of the underlying graph structure, machine learning models are frequently employed in social network analysis to infer user behavior, preferences, and social interactions. However, these models may inadvertently capture and encode sensitive information within their learned representations, such as private friendships, personal interests, and social circles \cite{li2020graph}.
This is also the case for the e-commerce domain, where social networks can be used as input to a recommender system~\cite{Campana_2017}.
Graph unlearning has the potential to provide means of mitigating this privacy concern. By employing graph unlearning techniques to the trained model, it is possible to delete or anonymize sensitive user data from the model's learned parameters. This promotes a more secure and ethical environment for social network analysis by ensuring that the model's predictions and insights are obtained without compromising the privacy and confidentiality of individual users. Through graph unlearning, researchers and organizations can strike a delicate equilibrium between leveraging the power of machine learning in understanding social dynamics and respecting and preserving users' privacy rights in the ever-changing social network landscape \cite{proferes2021studying,fuchs2011web}.

\subsection{Confidentiality in Financial Networks}

In financial transactions and fraud detection, learning models are used to analyze transaction patterns and identify anomalies. However, they may also capture sensitive information such as financial histories, transaction details, and account balances, jeopardizing user privacy. Graph unlearning can remove or obfuscate this data from learned representations, enabling accurate insights and fraud detection without retaining sensitive financial information \cite{glasserman2016contagion}. Adopting unlearning in financial networks provides a robust, ethical framework for privacy and security, fostering user confidence.

\subsection{Ethical Consideration in Healthcare Networks}

In healthcare networks, ethical considerations are of the uttermost importance. GML models can be used to analyze patient data, identify disease patterns, and support clinical decision-making in the healthcare industry. However, these models may inadvertently acquire confidential patient information, such as medical histories, diagnoses, and treatment plans, raising grave privacy concerns. These models may incorporate confidential health information within their learned representations, posing a substantial threat to patient confidentiality \cite{giota2014mental}. Integrating graph unlearning into healthcare networks may enables the effective protection of patient privacy.

\subsection{Ethical Consideration in Transportation Networks}

Commonly, machine learning models are employed in traffic analysis to predict traffic patterns, optimize routes, and enhance transportation systems \cite{derrow2021eta}. Such models may also capture sensitive details about individuals’ mobility patterns, commute routes, and location histories, creating privacy risks. Graph unlearning can remove this information from trained models, safeguarding traveler anonymity while maintaining the accuracy and efficiency of traffic predictions.

\subsection{Ethical Consideration in Biological Networks}

In the fields of bioinformatics and biological network analysis, machine learning models play a crucial role in the analysis of complex biomolecular interactions and functions. However, these models may accidentally discover intricate details about individual genetic sequences, gene expressions, behavioral information and metabolic pathways, which raises privacy concerns \cite{said2023neurograph}. The risk of data exposure and potential individual identification posed by these models can compromise the confidentiality and privacy of genomics data. By incorporating graph unlearning into biological networks, these ethical concerns can be effectively addressed.

\subsection{Individuals' Data Privacy in Collaborative Environment}

In collaborative research projects, for instance, researchers may be required to share learning models trained on sensitive data in order to reproduce and validate results; however, they must also safeguard the privacy of individual contributors. In this scenario, graph unlearning can enables researchers to share the model's insights without disclosing the underlying raw data, thereby protecting the privacy of data contributors. Similarly, in cross-organizational initiatives, businesses frequently collaborate to optimize operations and acquire a holistic understanding of their respective fields. However, they must protect the confidentiality of sensitive business information. In the context of smart city projects, in which various stakeholders seek to enhance urban services through data-driven strategies, there is an increasing demand to share data while protecting the privacy of citizens. Graph unlearning may enables city authorities to collaborate with private companies, researchers, and service providers without disclosing personal data, achieving a balance between urban development and data privacy. Moreover, in academic partnerships, researchers from various institutions may collaborate to resolve complex scientific problems, necessitating the exchange of data and models \cite{hu2011detecting}. Graph unlearning ensures that valuable research insights can be shared without violating data protection laws, thereby fostering productive academic collaborations.

\subsection{Adversarial Setting}

In sensitive domains like healthcare, finance, and personal recommendation systems, the consequences of a security breach compromising access to a trained machine learning model can be severe. In these scenarios, an attacker may attempt to jeopardize user privacy by injecting malicious data into the model's training dataset or by stealing sensitive information from the model itself. This malicious activity can lead to various detrimental outcomes, such as unauthorized access to personal medical records, financial data, or private user preferences \cite{olatunji2021releasing}. For instance, in the healthcare domain, if an attacker gains access to a medical diagnosis model, they could inject misleading or harmful data, potentially leading to inaccurate diagnoses and jeopardizing patient safety. In the finance sector, unauthorized access to a fraud detection model could enable attackers to manipulate the system, bypass security measures, and conduct fraudulent transactions. Similarly, in personal recommendation systems, adversaries may exploit information about users' preferences and behaviors obtained through malicious data injection or information stealing \cite{wu2022linkteller}. This could result in biased or manipulated recommendations, eroding user trust and leading to potential misuse of personal data. To safeguard against these risks, 
%
several recent methods have been proposed to mitigate the effects of adversarial attacks on GNNs. For instance, in \cite{wu2024graphmu}, the authors have introduced a three-step approach. First, attack detection methods are employed to identify anomalous nodes or edges within the graph. The accessibility of these anomalous nodes or edges varies depending on the specific scenario and the level of awareness regarding poisoned samples. Second, a fine-tuned subgraph is constructed based on the detected anomalous nodes or edges, ensuring that the subgraph is optimized for subsequent processing. Finally, the constructed subgraph is utilized to adjust or re-optimize the parameters of the poisoned GNN model, aiming to restore its predictive accuracy while mitigating the influence of adversarial perturbations.
In order to allow the utilization of new attack advancements, \cite{di2024adversarial} recently presented a game-theoretic approach that integrates membership inference attack (MIAs) into the framework of unlearning algorithms. In this setting, the unlearning task is formulated as a Stackelberg game, where the unlearner aims to remove specific training data from a model, while the auditor leverages MIAs to detect any residual evidence of the supposedly erased data.

\subsection{Limited Resource Environment}

The application of graph unlearning may holds immense potential, especially in limited resource environments such as the Internet of Things (IoT). In such scenarios, devices at the edge may encounter situations where certain training examples become obsolete, out of distribution, or no longer valid \cite{hassan2021leveraging,fan2022fast}. For instances, a network of smart traffic cameras placed strategically across the city to monitor traffic flow and detect any congestion or accidents.
These cameras usually share the same on-device model, which can be synchronously finetuned with federated learning as new data arrives. Moreover,
each traffic camera acts as a node in the knowledge graph, and the connections between these nodes represent the spatial relationships between different camera locations. Assume that we have different devices at different locations of the cities to collect data from the nearby cameras. As vehicles move through the city, the smart traffic cameras continuously capture real-time data, such as vehicle count, speed, and license plate recognition. This data is then used to generate and update the knowledge graph, which dynamically represents the traffic patterns and conditions in the city. Over time, the traffic patterns may change due to various factors, such as construction work, events, or changes in city planning. Some of the historical data captured by the traffic cameras on some devices may become less relevant for predicting and managing the current traffic conditions.
These observations incentivize the development of a graph unlearning method that, when applied to the distributed knowledge graph formed by the cameras, can
optimize the use of resources and improve the accuracy of traffic predictions.
Here, the concept of fine-tuning from an unlearning perspective becomes valuable, particularly when the devices aim to retain their existing data without adding more. Instead, these devices may selectively flag certain types of data for unlearning, allowing them to adapt to dynamic changes in their operating environment.
Furthermore, IoT devices have limited computational resources, which in turn gives a push to focus on unlearning efficiency. The common approach of periodically retraining all models become less feasible for the modern AI landscape, where large (foundation) models become more common, and edge devices such as smartphones become the norm. As a result, efficient graph unlearning algorithms would prove to be very beneficial as the field of ML keep progressing at a break neck pace.

\section{Discussion and Future Directions}
\label{sec:discussion}

Graph unlearning is a promising and crucial field with the potential to address data privacy and security concerns in the realm of graph-based machine learning models \cite{Zhang_2024}. As graph-based models continue to gain popularity for their capacity to capture intricate relationships and dependencies among data points in diverse domains, such as social network analysis, recommendation systems, and bioinformatics \cite{wu2020comprehensive}, the need for robust unlearning approaches becomes increasingly evident. The inherent nature of these models raises privacy issues, as they may accidentally store sensitive information about individuals or entities, jeopardizing user data confidentiality \cite{daigavane2021node}. In Section \ref{sec:categorization}, we presented a variety of graph unlearning approaches, including exact, and approximate unlearning, methods proposed in recent years. To build upon the existing literature on privacy-preserving approaches, the following sections include a discussion on differential privacy, accompanied by a brief review of relevant studies. Finally, potential research directions that warrant further exploration in future work will be proposed.

\subsection{Data Privacy}
\label{subsec:dpapproaches}

The potential leakage of sensitive information in scenarios where the trained model is shared or used in sensitive applications raises concerns about individual privacy \cite{daigavane2021node}. Differential Privacy (DP) offers solutions to this issue by introducing controlled noise into the model updates during SGD, ensuring that individual data points do not unduly influence the model's parameters. Well-known works in this line of research include \cite{abadi2016deep,dwork2008differential}, which introduces a novel algorithmic technique for training models under the umbrella of DP with tight privacy bounds. \cite{abadi2016deep} approach involves clipping the $l_2$ norm of each gradient, computing the average, adding noise to protect privacy, and then taking a step in the opposite direction of this average noisy gradient. Additionally, it incorporates a privacy loss function to further enhance privacy protection during training. Any DP approach can then be trivially applied to unlearning, as the former is a strictly stronger definition of the latter~\cite{sekhari2021rememberwantforgetalgorithms}.

Despite its desired theoretical guarantee, DP is not suitable for many cases given it requires careful prior planning. Solving this issue, certified removal methods aims to provide the same guarantee for when training was conducted without DP.
Beyond \cite{guo2020certified} that served as foundation for \cite{chien2022certified} to introduce certified unlearning on graphs, literature on this topic for general machine learning has moved forward while its graph-equivalent did not receive such interest. \cite{sekhari2021remember} similarly utilizes second-order updates to unlearn samples, but provides guarantee with respect to the test loss instead of the gradient norm.
\cite{zhang2025certifiedunlearningdeepneural} extends the previous ideas to work with non-convexity by proposing an efficient inverse Hessian approximation for the update.
Attacking from another angle, by embracing DP and noisily finetuning model on unforgotten data, \cite{koloskova2025certifiedunlearningneuralnetworks} utilizes gradient clipping and model clipping to circumvent the loss smoothness requirement, in turn bounding sensitivity and injected noise magnitude.
While certified removal is not as strong as exact unlearning, the certification could provide better credibility for weaker approximate unlearning algorithms through a theoretical guarantee; and thus exploring this underutilized area of research in the graph context may prove to be very valuable, especially in privacy-critical scenarios.


Naturally, developments in general ML privacy have been adapted to GNNs as they gain prominence.
Recent efforts apply DP to provide formal guarantees in various GNN scenarios. For edge-level privacy, \cite{wu2022linkteller} perturbs the input graph using randomized response (EdgeRand) or the Laplace mechanism (LapGraph) before training, though the method does not extend easily to node-level privacy. \cite{sajadmanesh2021locally} propose a locally private GNN in a federated setting where node features and labels remain private but edges are shared, limiting applicability when edges are sensitive. The GAP framework \cite{sajadmanesh2023gap} uses aggregation perturbation to protect edges, while \cite{olatunji2021releasing} adapt the PATE framework \cite{papernot2016semi} for node-level privacy, though reliance on public data constrains its use.

Privacy has also been addressed via federated and split learning. \cite{mei2019sgnn} combine structural similarity with FL to obscure content and structure, and \cite{jiang2020federated} develop a secure FL framework for video-derived graph sequences using secure aggregation. \cite{zhou2020privacy} split GNN computation across data holders with a trusted server, but dependence on such a party limits guarantees. Recognizing differing privacy needs for node attributes and structure, \cite{chien2023differentially} propose the GDP framework, introducing relaxed node-level data adjacency to balance topology and attribute privacy. \cite{dukler2023safe} present an alternative using shard graphs with lightweight cross-attention adapters \cite{dukler2023introspective}, linking each node to an adapter trained on its own and outward-edge data.

\subsection{Approximate Graph Unlearning for Non-Convex Settings}

While remarkable progress has been made in graph unlearning, a notable area that warrants further exploration is the development of approximate unlearning methods tailored to non-convex settings. Currently, many existing approaches primarily focus on convex settings, where optimization landscapes are well-behaved and lend themselves to more straightforward solutions \cite{pan2023unlearning,cong2023efficiently}. However, real-world graph-based machine learning applications often involve non-convex settings with complex and irregular optimization landscapes. As such, designing effective approximate unlearning techniques for these scenarios is challenging but crucial. The need for approximate unlearning in non-convex settings stems from the realization that many practical graph-based models operate in environments where convexity assumptions do not hold. Examples include MPNN and other highly expressive models, which are capable of capturing intricate patterns and dependencies within the data \cite{xu2018powerful,hamilton2017inductive}. In these settings, unlearning approaches need to cope with non-convex landscapes, which introduce additional intricacies making the task significantly more demanding \cite{bourtoule2021machine}.

Tackling approximate unlearning in non-convex settings is essential for the wider adoption of graph-based machine learning in various real-world applications \cite{hamilton2020graph}. The ability to efficiently remove sensitive information and adjust models in response to evolving privacy requirements will instill confidence among users and stakeholders in utilizing graph-based machine learning solutions. While this direction poses challenges due to the inherent difficulty of approximating models in non-convex settings, it presents an exciting opportunity to push the boundaries of graph unlearning research \cite{cheng2023gnndelete}. Solutions that can effectively handle non-convexity while maintaining a balance between privacy preservation and model performance will be invaluable in applications ranging from social network analysis, recommendation systems, to bioinformatics and beyond.

\subsection{Graph Unlearning Beyond Simple Graphs}

While the existing approaches have primarily focused on undirected graphs, many real-world scenarios involve more complex graph structures, such as directed, weighted, temporal, and knowledge graphs \cite{zhu2023heterogeneous}. Directed graphs, in particular, are prevalent in applications involving causality, flow of information, or directed relationships \cite{daigavane2021node}. Temporal graphs, on the other hand, capture time-varying relationships between entities, which are commonly found in dynamic systems and evolving networks \cite{said2023neurograph}. Extending unlearning techniques to address these diverse graph types opens up a plethora of new applications. For instance, knowledge graphs, which model relationships and semantic information between entities, play a critical role in knowledge representation and reasoning tasks. Unlearning in knowledge graphs can offer valuable insights into privacy-preserving knowledge-sharing environments, ensuring that sensitive information is suitably removed while retaining the utility of the underlying knowledge \cite{jiang2020federated}. Furthermore, directed and temporal graphs are prevalent in domains like social media analysis, financial transactions, and epidemiological studies, where understanding the directionality and evolution of relationships is essential. In these contexts, the ability to unlearn and protect sensitive information without compromising the temporal dynamics of the graph is of paramount importance \cite{wu2020comprehensive}. Addressing graph unlearning in directed and temporal graphs requires devising novel algorithms and strategies tailored to their unique characteristics. Considering that these graph types exhibit rich and intricate structures, the challenges lie in developing efficient unlearning techniques that take into account causality, directed information flow, and temporal dependencies.

This research direction has the potential to revolutionize privacy and security measures in various graph-based machine learning applications. By enabling the unlearning of sensitive information in directed, weighted, temporal, and knowledge graphs, data privacy can be preserved while facilitating the development of more privacy-aware and responsible AI systems. The impact of such advancements is far-reaching, with applications spanning social network analysis, recommendation systems, epidemiology, finance, and more \cite{derrow2021eta}.

\subsection{Universal Graph Unlearning Approaches}

Universal graph unlearning represents a transformative direction in the field of graph unlearning, encompassing the capability to perform any unlearning task, including node, edge, feature unlearning, and graph-level unlearning. While the current approaches have made significant progress in addressing specific unlearning tasks, they often remain limited, specializing in one or a few types of unlearning tasks. The need to focus on designing universal approaches that can execute all unlearning tasks is paramount, considering that a majority of GNNs are designed to handle a wide range of tasks \cite{kipf2016semi}. The versatility of universal graph unlearning offers profound benefits, as it caters to the diverse requirements of various graph-based machine learning applications. Different unlearning tasks address distinct aspects of data privacy and security in graphs. For instance, node unlearning is essential for preserving individual-level information, edge unlearning addresses the removal of specific relationships, feature unlearning ensures the protection of sensitive node attributes, and graph-level unlearning deals with privacy at a holistic level.

Designing universal approaches is challenging, as it necessitates accommodating the intricacies of multiple unlearning tasks within a single framework. Moreover, ensuring the efficiency and scalability of universal graph unlearning is vital to enable its application on large-scale graph datasets and complex graph structures \cite{pan2022unlearning}. By embracing the concept of universal graph unlearning, researchers can create a unified framework that seamlessly integrates with graph-based machine learning models, including GNNs with all graph unlearning tasks. This will foster the development of more comprehensive and privacy-aware AI systems, where data privacy can be adequately preserved across various unlearning tasks. Universal graph unlearning holds the potential to become a fundamental tool in enhancing data privacy and security measures, offering a holistic approach to address privacy concerns in graph-based machine learning.

\subsection{Leveraging Graph Descriptors for Graph-level Unlearning}

Another promising research direction is the exploration of graph descriptors or graph embedding methods when focusing solely on graph-level unlearning. Graph descriptors, also known as graph embeddings, represent graphs as vectors or low-dimensional representations in a continuous space \cite{cai2018comprehensive}. They capture the structural information and relationships within the graph, enabling various machine learning algorithms to operate on graphs effectively. A significant advantage of using graph descriptors for graph-level unlearning is that once a graph embedding is computed, it remains fixed and unaffected by subsequent unlearning operations, as no further actions are needed. Moreover, majority of graph descriptors are extremely efficient to compute \cite{tsitsulin2018netlsd,index2021netki}. This characteristic makes graph descriptors particularly well-suited for scenarios where frequent model updates or unlearning procedures may or may not be desirable.

Several graph descriptor methods have emerged in recent years, each with its unique characteristics. For further reading, we refer the reader to recent works on graph descriptors \cite{said2021dgsd, index2021netki,cai2018comprehensive}. By incorporating graph descriptors into graph-level unlearning, researchers can take advantage of the rich representations they offer, allowing for efficient unlearning without the need to retrain the model repeatedly. The fixed nature of graph embeddings ensures that graph-level privacy-preserving operations do not interfere with the overall performance and generalization capabilities of the machine learning model. Exploring novel graph descriptor techniques tailored to graph-level unlearning requirements is an exciting avenue for future research. Additionally, comparative studies that evaluate different graph embedding methods in the context of unlearning tasks will provide valuable insights into their strengths and limitations. Moreover, since these descriptors are quite efficient, new embeddings could be quickly computed if node or edge-level removal requests are received.

\subsection{Graph Unlearning for User-to-User Interactions}
The right to be forgotten is straightforward when only a single user is involved, or when the entities directly affected are items (e.g., in recommendation systems). However, in user-interaction-driven scenarios where the interactions are among users, more discussions are necessary. For example, on online dating platforms, if one user requests unlearning to delete a record, is it fair to the other user involved in that interaction? Should they have the right to retain the memory of the event? These scenarios, where interactions occur among users, present unique challenges in addressing unlearning, requiring careful consideration of the rights of all involved parties.

\begin{figure}
    \centering
    \includegraphics[width=0.9\linewidth]{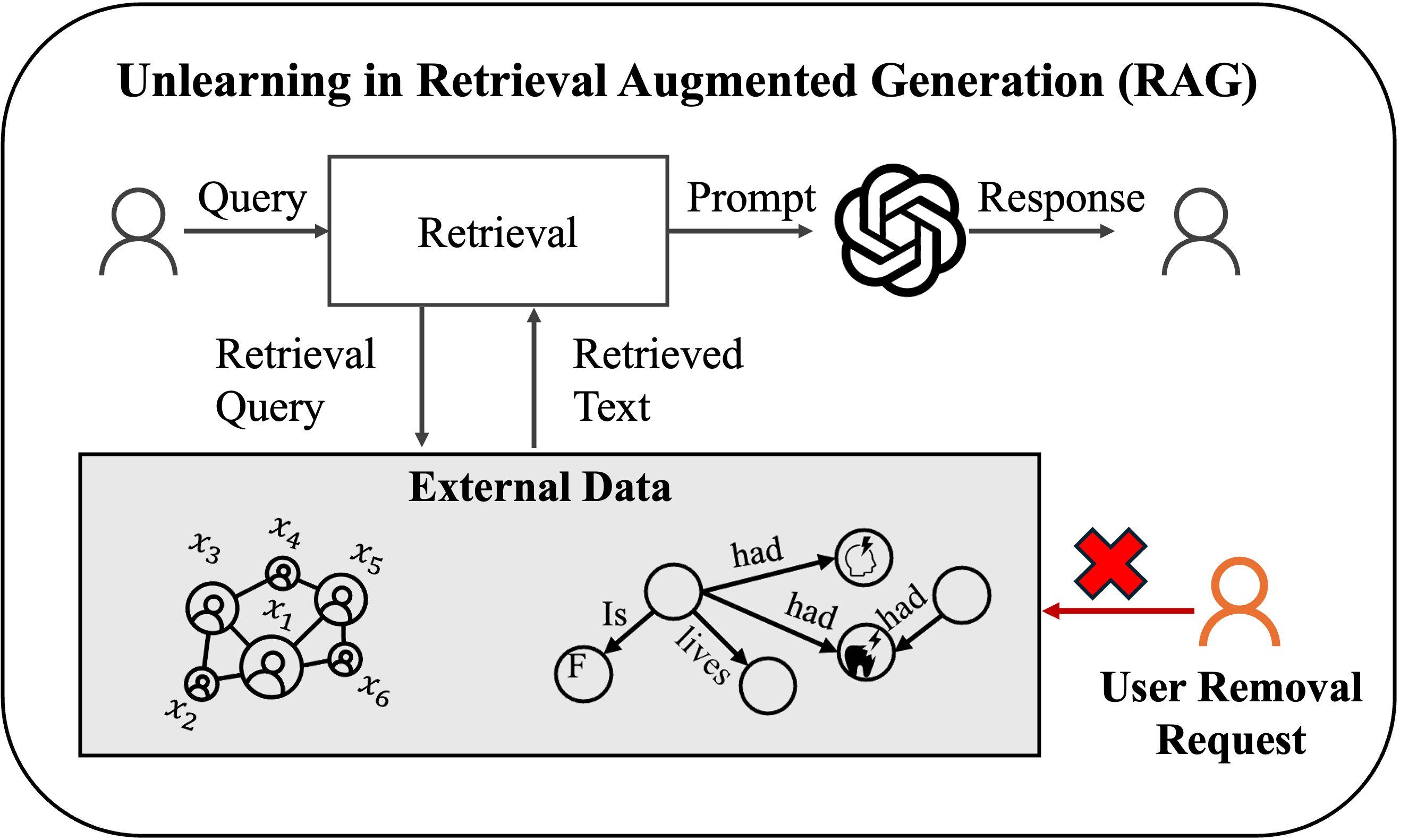}
    \caption{Unlearning in Retrieval Augmented Generation}
    \label{fig:rag}
    \vspace{-3mm}
\end{figure}

\subsection{Unlearning for Foundation Models}
With the success of large language models (LLMs) \cite{zhao2023survey}, large-scale foundation models have been developed across various domains, including graphs \cite{mao2024graph,wang2024large}. These models typically follow a common paradigm of pre-training followed by fine-tuning. However, despite advancements in efficient unlearning algorithms, directly unlearning from the pre-trained model remains time-consuming due to its large size. This highlights the need to explore unlearning during the fine-tuning stage. Retrieval-augmented generation (RAG) has been commonly employed to enhance the performance of foundation models, but RAG relies on external datasets that might contain data a user seeks to remove, as seen in Figure~\ref{fig:rag}. Many studies leverage knowledge graphs as the external data source \cite{he2024g, peng2024graph} due to the powerful representation capabilities of graphs, making the challenges of graph unlearning unavoidable in these graph-based RAG systems.
More recently, a new LLM-inspired approach to finetune foundation graph model has been graph prompt tuning~\cite{gppt,fang2024universalprompttuninggraph}, where most of the model is frozen with the exception of a small subset of weights that directly modify the input graph itself---these weights are then updated to work with the target downstream task.
This naturally creates the demand for adapting existing unlearning methods for LLMs~\cite{bhaila2024softpromptingunlearninglarge,chowdhury2025scalableexactmachineunlearning} to the graph setting, and analyzing whether existing pitfalls for LLMs~\cite{wei2025llms} persist to the graph foundation models, presents an intriguing research direction. 

\section{Conclusion}

This paper offers a comprehensive review of the emerging field of graph unlearning, which plays a vital role in responsible AI development. It details various graph unlearning approaches and methodologies, along with their applications in diverse domains, highlights the wide-ranging impact of this research area, including social networks, financial networks, and transportation systems. We emphasize the sensitivity of GML to data privacy and adversarial attacks, underscoring the importance of graph unlearning techniques in addressing these critical concerns.
The comprehensive taxonomy, up-to-date literature overview, and evaluation measures provided in this survey empower researchers to explore new approaches and contribute to this evolving field. The suggested research directions offer exciting opportunities for future investigations, encouraging the development of novel algorithms, fortified privacy guarantees, and exploration of applications in emerging domains.

\bibliographystyle{ieeetr}
\bibliography{arXiv_upload/main}

\end{document}